\documentclass[10pt,twocolumn]{article}
\usepackage[super,numbers]{natbib}
\usepackage[utf8]{inputenc}
\usepackage{graphicx}
\usepackage[colorlinks,citecolor=blue,linkcolor= blue,urlcolor=blue]{hyperref}
\usepackage[affil-it]{authblk}
\usepackage{mathptmx}
\usepackage{enumitem}
\usepackage[font=small]{caption}

\usepackage[compact]{titlesec}
\titleformat*{\section}{\normalsize\bfseries}
\titleformat*{\subsection}{\normalsize\bfseries}
\titleformat*{\subsubsection}{\normalsize\bfseries}
\titleformat*{\paragraph}{\normalsize\bfseries}
\titleformat*{\subparagraph}{\normalsize\bfseries}
\titlespacing*{\section}{0pt}{2.5ex plus 0ex minus 0ex}{0ex plus 0ex minus 0ex}
\titlespacing*{\subsection}{0pt}{2.5ex plus 0ex minus 0ex}{0ex plus 0ex minus 0ex}
\titlespacing*{\subsubsection}{0pt}{2.5ex plus 0ex minus 0ex}{0ex plus 0ex minus 0ex}

\usepackage{geometry}
\title{\raggedright\vspace{-1.3cm}\Huge\textbf{Empowering Local Communities\\Using Artificial Intelligence}}
\author{
	\raggedright
    Yen-Chia Hsu\textsuperscript{\rm 1,*}, Ting-Hao `Kenneth' Huang\textsuperscript{\rm 2}, Himanshu Verma\textsuperscript{\rm 1}, Andrea Mauri\textsuperscript{\rm 1,\rm 4}, Illah Nourbakhsh\textsuperscript{\rm 3}, Alessandro Bozzon\textsuperscript{\rm 1}
}
\affil{
	\vspace{-0.1cm}
	\raggedright
	\normalsize
    \textsuperscript{\rm 1}\normalfont Faculty of Industrial Design Engineering, Delft University of Technology, Delft, Netherlands \\
    \textsuperscript{\rm 2}\normalfont College of Information Sciences and Technology, Pennsylvania State University, State College, PA, USA \\
    \textsuperscript{\rm 3}\normalfont Robotics Institute, Carnegie Mellon University, Pittsburgh, PA, USA \\
    \textsuperscript{\rm 4}\normalfont Amsterdam Institute for Advanced Metropolitan Solutions, Amsterdam, Netherlands \\
    \textsuperscript{*}\normalfont Corresponding Author and Lead Contact: y.hsu-1@tudelft.nl
}
\date{\vspace{-1.1cm}}

\begin{document}
\twocolumn[
\begin{@twocolumnfalse}
	\maketitle
	\setlength{\fboxsep}{13pt}
	\fbox{\begin{minipage}{1.95\columnwidth}
		\paragraph*{\small THE BIGGER PICTURE}
		Artificial Intelligence (AI) is increasingly used to analyze large amounts of data in various practices, such as object recognition. We are specifically interested in using AI-powered systems to engage local communities in developing plans or solutions for pressing societal and environmental concerns. Such local contexts often involve multiple stakeholders with different and even contradictory agendas, resulting in mismatched expectations of these systems' behaviors and desired outcomes. There is a need to investigate if AI models and pipelines can work as expected in different contexts through co-creation and field deployment. Based on case studies in co-creating AI-powered systems with local people, we explain challenges that require more attention and provide viable paths to bridge AI research with citizen needs. We advocate for developing new collaboration approaches and mindsets that are needed to co-create AI-powered systems in multi-stakeholder contexts to address local concerns.
	\end{minipage}}
	\vspace{0.5cm}
\end{@twocolumnfalse}
]

\section*{SUMMARY}

Artificial Intelligence (AI) applications can profoundly impact society. Recently, there has been extensive interest in studying how scientists design AI systems for general tasks. However, it remains an open question about whether the AI systems developed in this way can work as expected in different regional contexts while simultaneously empowering local people. How can scientists co-create AI systems with local communities to address regional concerns? This article contributes new perspectives in this under-explored direction at the intersection of data science, AI, citizen science, and human-computer interaction. Through case studies, we discuss challenges in co-designing AI systems with local people, collecting and explaining community data using AI, and adapting AI systems to long-term social change. We also consolidate insights into bridging AI research and citizen needs, including evaluating the social impact of AI, curating community datasets for AI development, and building AI pipelines to explain data patterns to laypeople.

\section*{INTRODUCTION}

Artificial Intelligence (AI) techniques are typically engineered with the goals of high performance and accuracy. Recently, AI algorithms have been integrated into diverse and real-world applications. It has become an important topic to explore the impact of AI on society from a people-centered perspective~\cite{shneiderman2020bridging}. Previous works in citizen science have identified methods of using AI to engage the public in research, such as sustaining participation, verifying data quality, classifying and labeling objects, predicting user interests, and explaining data patterns~\cite{franzen2021machine,ceccaroni2019opportunities,lotfian2021partnership,mcclure2020artificial}. These works investigated the challenges regarding how scientists design AI systems for citizens to participate in research projects at a large geographic scale in a generalizable way, such as building applications for citizens globally to participate in completing tasks. In contrast, we are interested in another area that receives significantly less attention:
\begin{itemize}[noitemsep,topsep=0pt]
	\item How can scientists co-create AI systems ``with'' local communities to address context-specific concerns and influence a particular geographical region?
\end{itemize}
Our perspective is based on applying AI in Community Citizen Science~\cite{hsu2020human,chari2017promise} (CCS), a framework to create social impact through community empowerment at an intensely place-based local scale. We define ``community'' as a group of people who are indirectly or directly affected by issues in civil society and are dedicated to making sure that these issues are recognized and resolved. We define ``social impact'' as how a project influences the society and local communities that are affected by social or environmental issues. We define ``community empowerment'' as a process of yielding agency to communities so that they can use technology, data, and informed rhetoric to create and disseminate evidence to advocate for social and policy changes. The CCS framework, a branch of citizen science~\cite{shirk2012public,irwin2001constructing}, is beneficial in co-creating solutions and driving social impact with communities that pursue the Sustainable Development Goals~\cite{fritz2019citizen}. Based on the literature and our experiences in co-creating AI systems with citizens, this article provides critical reflections regarding this under-explored topic for data science, AI, citizen science, and human-computer interaction fields. We discuss challenges and insights in connecting AI research closely to social issues and citizen needs, using prior works as examples.

\subsection*{How CCS Links to Other Frameworks}

Community Citizen Science emphasizes close collaborations among stakeholders when tackling local concerns. It is inspired by community-based participatory research~\cite{wallerstein2006using} and popular epidemiology~\cite{brown1993public}, where citizens directly engage in gathering data and extracting knowledge from these data for advocacy and activism. Examples involve co-designing technology for local watershed management~\cite{preece2019interaction}, understanding water quality with local communities~\cite{carroll2019empowering,jollymore2017citizen}, and using geo-information tools to monitor noise and earthquakes~\cite{carton2017citizen}. CCS intends to extend previous frameworks' scope to Sustainable Development Goals, especially the goal of sustainable cities and communities. This article discusses using CCS to integrate AI in-the-wild and local regions, which is different from those that conducted studies in living lab environments (such as the work by~\citet{alavi2018hide}) or in online communities (such as the work by~\citet{brambilla2014community}).

Additionally, Community Citizen Science is related to Action Research~\cite{susman1978assessment}, RtD (research through design)~\cite{zimmerman2007research}, Service Design~\cite{zomerdijk2010service}, and the PACT framework (participatory approach to enable capabilities in communities)~\cite{bondi2021envisioning}. Extending Action Research, CCS encourages scientists to immerse themselves in the field by taking on a social role and conducting research from a first-person view. Complementing RtD that creates prototypes as proof-of-concept, CCS develops functional systems that can be deployed and used by local people. Unlike Service Design, citizens’ roles extend beyond service consumers to co-designers who co-create knowledge and systems with scientists and other stakeholders. The PACT framework and CCS share the same goal of co-designing AI systems to address critical societal issues, while CCS has an additional goal that needs to be achieved simultaneously: empowering local communities to catalyze social impact.

\begin{figure*}[t]
	\centering
	\includegraphics[width=2.05\columnwidth]{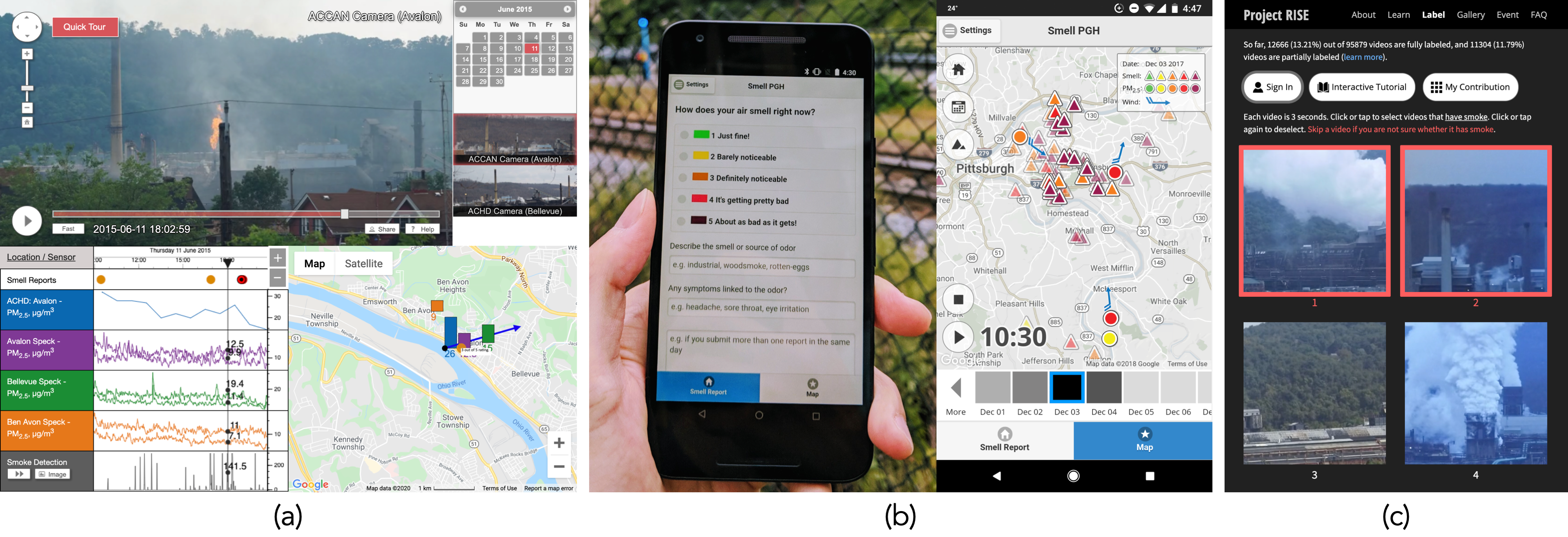}
	\vspace{-0.1cm}
	\caption{Case studies of Community Citizen Science projects that involve co-designing AI tools with local communities: (a) the air pollution monitoring project~\cite{hsu2017community} that empowered the Pittsburgh community to collect air pollution evidence in the local region for taking action, (b) the Smell Pittsburgh project~\cite{hsu2020smell} that invites citizens to report pollution odors and use the data as evidence to conduct air pollution studies, (c) the RISE project~\cite{hsu2021project} that enables citizens and scientists to annotate industrial smoke emissions and build an AI model to recognize pollution events. These cases were approved by the ethical committee of the university that hosted the projects.}
	\label{fig:prior_work}
\end{figure*}

\section*{CHALLENGES}

Due to its region-based characteristics, Community Citizen Science often involves many local stakeholders—including communities, citizens, scientists, designers, and policy-makers—with complex relationships. CCS creates the space for the stakeholders to reveal underlying difficulties and locally-oriented action plans in tackling social concerns that are hard to uncover in traditional technology-oriented and researcher-centered approaches. But, stakeholders often have divergent and even contradicting values, which results in conflicts that pose challenges when designing, engineering, deploying, and evaluating AI systems.

Based on case studies (Figure~\ref{fig:prior_work}) of co-creating AI systems with local people, we outline three major challenges:
\begin{itemize}[noitemsep,topsep=0pt]
	\item Co-designing AI systems with local communities
	\item Collecting and explaining community data using AI
	\item Adapting AI systems to long-term social changes
\end{itemize}
These challenges come mainly from the conflicts of interest between local communities (e.g., citizens and community groups) and university researchers (e.g., designers and scientists). AI researchers pursue knowledge to advance science, while local communities often desire social change. Such conflicts of interest among these two groups can lead to tensions, socio-technical gaps, and mismatched expectations when co-designing and engineering AI systems. For instance, local communities need functional and reliable systems to collect evidence, but AI researchers may be interested in producing system prototypes only to prove concepts or answer their research questions. Local community concerns can be urgent and timely, and citizens need to take practical actions that can have an immediate and effective social impact, such as public policy change. But, scientists need to produce knowledge using rigorous methods and publish papers in the academic community, often requiring a long reviewing and publication cycle.

\subsection*{Co-Designing AI Systems with Local Communities}

Challenges exist in community co-design, especially when translating multi-faceted community needs to implementable AI system features without using research-centered methods. The current practice to design AI systems is mainly centered on researchers instead of local people. Popular research methods—such as participatory design workshops, interviews, surveys—are normally used to help designers and scientists understand research questions. Although these methods enable researchers to better control the research process, essentially, university researchers are privileged and in charge of the conversations, leading to inappropriate power dynamics that can hinder trust and exacerbate inequality~\cite{klein2011dismantling,harrington2019deconstructing}. For example, during our informal conversations with local people that suffer from environmental concerns in our air quality monitoring project~\cite{hsu2017community}, many expressed feelings that scientists often treated them as experimental subjects (but not collaborators) in research studies. This imbalanced power relationship leads to difficulties in initiating conversations with citizens during our community outreach efforts.

Also, community data and knowledge are hyperlocal, which indicates that their underlying meanings ground closely to the local region and can be difficult to grasp for researchers who are not a part of the local community~\cite{carroll2015reviving}. For example, citizen-organized meetings to discuss community actions are often dynamic and context-dependent, which is not designed nor structured for research purposes. To collect research data that represents community knowledge, the current intensive procedure, such as video or audio recording, can make citizens feel uncomfortable. One alternative is to be a part of the community, to design solutions with them, join their actions, and perform ethnographic observations. For instance, researchers can better understand local community needs by actively participating in regional citizen group meetings and daily conversations with citizens. However, such in-depth community outreach approaches take tremendous personal effort, which can be unmotivating or even infeasible due to the limited academic research cycle and research-oriented academic tenure awarding system~\cite{klein2011dismantling,wallerstein2006using}.

\subsection*{Collecting and Explaining Community Data Using AI}

Challenges arise in data collection, analysis, and interpretation also due to conflicts of interest among scientists and citizens. Scientists are looking for rigorous procedures, but citizens seek evidence for action. Local communities are often frustrated by the formal scientific research procedure to prove the adverse impact of risk~\cite{brown1993public}, such as finding evidence of how pollution negatively affects health. Traditional environmental risk assessment models require a causal link between the risk and the outcome with statistical significance before taking action, which can be very difficult to achieve due to complex relationships between local people and their environments~\cite{bidwell2009community}. As a result, citizens collect their own community data (as defined by~\citet{carroll2018strengthening}, such as smoke emission photos from a nearby factory) as an alternative in order to prove their hypotheses. But, from scientists’ point of view, such strong assumption-driven evidentiary collection can lead to biases since the collection, annotation, and analysis of community data are conducted in a manner that strongly favors the assumption. One example is confirmation bias, where citizens are incentivized to search for information and provide data that confirms their prior beliefs~\cite{nickerson1998confirmation}, such as a high tendency to report odors related to pollution events~\cite{hsu2020smell}. Based on our experiences, it is extremely difficult to address or eliminate such biases when analyzing and interpreting how local social or environmental concerns affect communities.

Furthermore, researchers need to evaluate the social impact of AI systems to understand if the community co-design approach is practical. However, it is hard to determine if the intervention of AI systems actually influences the local people and leads to social changes by statistically analyzing community data. One difficulty is that local communities may have the implicit cognitive bias to overestimate and overstate the effect of the intervention since they are deeply involved in the co-creation of the AI system~\cite{norton2012ikea}. Moreover, it can be infeasible to conduct randomized experiments to prove the effectiveness of the intervention of AI on local communities~\cite{hsu2020human}. Controlling volunteer demographics and participation levels can be unethical when analyzing impact among different groups of people. Community Citizen Science treats local people as collaborators rather than participants. Researchers in CCS take the supporting role to assist communities using technology, instead of supervising and overseeing the entire project~\cite{hsu2020human}. Therefore, citizens join the CCS project at will and are not recruited like typical research studies. AI systems, in this case, are deployed in the wild with real consequences rather than a controlled test-bed environment that is designed for hypothesis testing. It remains an open research question as to how to integrate social science when studying the impact of AI systems~\cite{sloane2019ai}.

\subsection*{Adapting AI Systems to Long-Term Social Changes}

Conflicts of interest in the diverged values of citizens and scientists can lead to challenges in adapting AI systems to long-term social changes. The relationship between local people and AI systems is a feedback loop, which is similar to the concept that human interactions with architectural infrastructure are a continuous adaptation process that spans over long periods of time~\cite{alavi2019temporality}. When embedded in the social context, AI systems interact with citizens daily as community infrastructure. Communities are dynamic and frequently evolve their agenda to adapt to the social context changes. This means that the AI systems also need to adjust to such changes in local communities continuously. For instance, as we understand more about the real-life effects of the deployed AI systems on local people, we may need to fine-tune the underlying machine learning model using local community data. We may also need to improve the data analysis pipeline and strategies for interpreting results to fit local community needs in taking action. We may even need to stop the AI system from intervening in the local community under certain conditions. Such adaptation at scale requires ongoing commitment from researchers, designers, and developers to continuously maintain the infrastructure, involve local people in assessing the impact of AI, adjust the behavior of AI systems, and support communities in taking action to advocate for social changes~\cite{sloan2020participation}.

However, it is very challenging to estimate and obtain the required resources to sustain such long-term university-community engagement with local people~\cite{koekkoek2021unraveling}, especially in financially supporting local community members for their efforts. Typical research procedures can be laborious in data collection and analysis, and engineering AI systems with local people requires tremendous community outreach effort to establish mutual trust. In our experiences, applying and evaluating AI in Community Citizen Science relies heavily on an environment that has a sustainable fundraising mechanism in community organizations and universities. For example, funding is needed to hire software engineers that can maintain AI systems as community infrastructure in the long term, which can be hard to achieve in the current academic grant instruments and funding cycles.

The success of Community Citizen Science also depends on sustainable participation, which requires high levels of altruism, high awareness of local issues, and sufficient self-efficacy among local people. But, the complexity of the underlying machine learning techniques can affect the willingness to participate. On one side is whether the automation technique is trustworthy. In our experiences, local communities often perceive AI as a mysterious box that can be questionable and is not always guaranteed to work. Hence, citizens’ willingness to provide data can be low, but AI systems that employ machine learning and computer vision need data to be functional. On the other side, ``what citizens think the AI system can do'' does not match ``what the AI system can actually do'', resulting in socio-technical gaps and pitfalls for actual usage~\cite{roberts2021common}. In our experiences, local communities often have high expectations about what AI techniques can do for them, for example, automatically determining if an industrial site is violating environmental regulations. However, in practice, the AI system may only identify whether a factory emits smoke and degrades the air quality through sensors and cameras, which requires additional human efforts to verify if the pollution event is indeed a violation.

\section*{BRIDGE AI RESEARCH AND CITIZEN NEEDS}

University researchers typically lead the development of AI systems using a researcher-centered approach, where they often have more power over local communities (especially underserved ones) in terms of scientific authority and available resources. This unequal power relationship can result in a lack of trust and cause harm to underserved communities~\cite{koekkoek2021unraveling,harrington2019deconstructing}. An underlying assumption of this researcher-centered approach is that designers and scientists can put themselves in the situation of citizens and empathize with local people’s perspectives. However, university researchers are in a privileged situation in terms of socio-economic status and may come from other geographical regions or cultures, which means it can be very challenging for researchers to understand local people’s experiences fully and authentically~\cite{brown1993public}. Only by admitting this weakness and recognizing the power inequality can researchers truly respect community knowledge and be sincerely open-minded in involving local communities—especially those impacted by the problems the most—in the center of the design process when creating AI systems. Beyond being ``like'' the local people and designing solutions ``for'' them, researchers need to be ``with'' people who are affected by local concerns to co-create historicity and ensure that the AI systems are created to be valuable and beneficial to them~\cite{bennett2019promise,sloan2020participation}.

The critical role of creating social impact lies in local people and their long-term perseverance in advocating for changes. We believe that scientists need to collaborate with local people to address pressing social concerns genuinely, and even further, to immerse themselves into the local context and become citizens, hence ``scientific citizens'' (as defined by~\citet{irwin2001constructing}). However, pursuing academic research and addressing citizen concerns require different (even contradicting) efforts and can be difficult to achieve at the same time. Academic research requires contributing papers with scientific knowledge primarily to the research community, while citizen concerns typically involve many other stakeholders in a large and regional socio-technical system. It remains an open question how scientists and citizens can collaborate effectively under such dynamic, hyperlocal, and place-based conditions~\cite{bozzon2015needs}.

To move forward, we propose three viable Community Citizen Science approaches about how AI designers and scientists can conduct research and co-create social impact with local communities:
\begin{itemize}[noitemsep,topsep=0pt]
	\item Evaluate AI's social impact as empirical contributions
	\item Curate community data as dataset contributions
	\item Build AI pipelines as methodological contributions
\end{itemize}
These approaches produce empirical, dataset, and methodological contributions respectively to the research community, as defined by~\citet{wobbrock2016research}. To the local people, these approaches establish a long-term fair university-community partnership in addressing community concerns, increase literacy in collecting community data, and equip communities with AI tools to interpret data. CCS projects will succeed when designers and scientists see themselves as citizens, and in turn, when local communities and citizens see themselves as innovators. It is essential for all parties to collaborate around the lived experiences of one another and listen to each other’s voices with humility and respect.

\subsection*{AI's Social Impact as Empirical Contributions}

Lessons learned from previously deployed AI systems in other contexts cannot be simply applied in the current one, as local communities have various cultures, behaviors, beliefs, values, and characteristics~\cite{sloane2019ai}. Hence, it is essential to understand and document how scientists can co-design AI systems with local communities and co-create long-term social impact in diverse contexts. It is also important to study the effectiveness and impact of various AI interventions with different design criteria in sustaining participation, affecting community attitude, and empowering people. The Community Citizen Science framework provides a promising path toward these goals. Implications of collaborating with local people in co-designing AI interventions, creating long-term impacts, and tackling the conflicts of interest among stakeholders can be strong empirical contributions to the academic community~\cite{sloan2020participation}. The data-driven evidence and the interventions that are produced by AI systems can impact the local region in various ways, including increasing residents’ confidence in addressing concerns, providing convincing evidence, or rebalancing power relationships among stakeholders.

For instance, our air pollution monitoring project documented the co-design process regarding how designers translated citizen needs and local knowledge into implementable AI system features, as recognized by the Human-Computer Interaction community and published on ACM CHI~\cite{hsu2017community}. This work shows researchers how we co-created an AI system to support citizens in collecting air pollution evidence and how local communities used the evidence to take action. For example, in the computer vision model for finding industrial smoke emissions in videos, the feature vectors are handcrafted according to the behaviors and characteristics of smoke, which are provided by community knowledge. Also, the communities decide the areas in the video that require image processing. The decision of having high precision in the prediction (instead of high recall) is also a design choice by local people for quickly determining severe environmental violations. Another example is our study of push notifications which are generated by an AI model to predict the presence of bad odor in the city~\cite{hsu2020smell}. The finding from the study explains how sending certain types of push notifications to local citizens is related to the increase of their engagement level, such as contributing more smell reports or browsing more data.

Although one may not simply duplicate the collaboration ecosystem in these contexts due to unique characteristics in the local communities, our projects can be seen as case studies in specific settings. Our air quality monitoring case provides insights to researchers working on similar problems in other contexts about integrating technology reliably into their settings, as cited by~\citet{ottinger2017crowdsourcing}. Moreover, the case also helped researchers understand and categorize different modes of community empowerment, as cited by~\citet{schneider2018empowerment}.

\subsection*{Community Data as Dataset Contributions}

Data work is critical in building and maintaining AI systems~\cite{sambasivan2021everyone}, as modern AI models are powered by large and constantly changing datasets. When addressing local concerns with the support of AI systems, researchers often need to finetune existing models or build new pipelines to fit local needs. This requires collecting data in a specific regional context and may introduce new tasks to the AI research field. Community Citizen Science provides a sustainable way to co-create high-quality regional datasets while simultaneously increasing citizens’ self-efficacy in addressing local problems. Based on our experiences, co-creating publicly available community data can also facilitate citizens’ sense of ownership of the collaborative work. Such value of community empowerment links AI research closely to social impact and public good.

Besides the value of increasing citizens’ data literacy, the collected real-world data, the data collection approach, and the data processing pipeline can be combined into a significant dataset contribution to the academic community in creating robust AI models. Such community datasets are gathered in the wild with local populations over a long-term period to reflect the regional context, which complements the datasets obtained using crowdsourcing approaches (such as Amazon Mechanical Turk) in a broader context. In this way, community datasets provide values for AI researchers to validate if AI models trained on general datasets can work as expected in different regional contexts. Also, the accompanying software for data labeling can contribute reusable computational tools to the research community that investigates data annotation strategies.

For example, our RISE project presented a novel video dataset for the smoke recognition task, which can help other researchers develop better computer vision models for similar tasks, as recognized by the Artificial Intelligence community and published on AAAI~\cite{hsu2021project}. Our project demonstrated the approach of collaborating with citizens affected by air pollution to annotate videos with industrial smoke emissions at large scale under various weather and lighting conditions. The dataset was used to train a computer vision model to recognize industrial smoke emissions, which allowed community activists to curate a list of severe pollution events as evidence to conduct studies and advocate for enforcement action. Another example is the Mosquito Alert project that curates and labels a large mosquito image dataset with local people using a mobile application~\cite{pataki2021deep}. The dataset is built with local community knowledge and is used to train a mosquito recognition model to support the local public health agency in disease management. Besides its social impact, the Mosquito Alert project advances science by providing a real-world dataset for researching different mosquito recognition models, as cited by~\citet{adhane2021deep}.

\subsection*{AI Pipelines as Methodological Contributions}

In Community Citizen Science, there is a need to unite expertise from the local communities and scientists to build AI pipelines using machine learning to assist data labeling, predict trends, or interpret patterns in the data. An example is to forecast pollution and find evidence of how pollution affects the living quality in a local region. Although the concept of machine learning is common among computer scientists, it can look mysterious to citizens.

Thus, during public communication and community outreach, researchers often need to visualize analysis results and explain the statistical evidence for local residents, which is highly related to the Explainable AI (XAI) and interpretable machine learning research~\cite{doshi2017towards}. However, current XAI research mainly focuses on making AI understandable for experts rather than laypeople and local communities~\cite{cheng2019explaining}. This creates a unique research opportunity to study co-design methods and software engineering workflows of translating AI models’ predictions and their internal decision-making process into human-intelligible insights in the hyperlocal context~\cite{shneiderman2020bridging,miller2019explanation,burrell2016machine}. We believe the pipeline of such translation into explainable evidence can be a methodological contribution to the academic community, which provides a way to deal with the challenge of predicting future trends and interpreting similar types of real-world data. Also, the implemented machine learning pipeline and the design insights of developing the pipeline can contribute reusable computational tools and novel software engineering workflows to the research communities that study Explainable AI and its user interfaces.

For instance, our Smell Pittsburgh project used machine learning to explain relationships between citizen-contributed odor reports and air quality sensor measurements, contributing to a methodological pipeline of translating AI predictions, as recognized by the Intelligent User Interface community and published on ACM TiiS~\cite{hsu2020smell}. In this way, the pollution patterns became visible for public scrutiny and debate. Another example is the xAire project that co-designed solutions with local schools and communities to collect Nitrogen Dioxide data~\cite{perello2021large}. The air measurements were analyzed with asthma cases in children using a statistical machine learning model. The community outreach and public communication enabled laypeople to make sense of how Nitrogen Dioxide posed a risk to local community health. The pipelines in these two examples produced meaningful patterns for citizens to understand and communicate about how pollution impacts the local region. They also informed researchers about how to process, wrangle, analyze, and interpret urban data in order to explain insights to laypeople.

\section*{NEXT STEPS}

We have explained major challenges in co-designing AI systems with local people and empowering them to create broader social impact. We also proposed Community Citizen Science approaches to simultaneously addressing local societal issues and advancing science. Computing research communities have made steps to recognize the impact of technology and AI on society, such as establishing a separate track for paper evaluation (e.g., the AAAI Special Track on AI for Social Impact). We urge the computing research communities to go further and acknowledge social impact as a type of formal contribution in scientific inquiry and paper publication. Promoting this kind of contribution can be a turning point to encourage scientists to link research to society and ultimately make university research socially responsible for the public good. Co-creating AI systems and developing reusable tools with local communities in the long term allows scientists and designers to explore real-world challenges and solution spaces for various AI techniques, including machine learning, computer vision, and natural language processing.

We also urge universities to integrate social impact into the evaluation criteria of the tenure roadmap of the academic professorship as the ``service'' pillar of the university that contributes to the public good. We envision that applying Community Citizen Science when co-designing AI systems can advance science, build public trust in AI research through genuine reciprocal university-community partnership, and directly support community action to impact society. In this way, we may fundamentally change how universities, organizations, and companies partner with their neighbors to pursue shared prosperity in the future of Community-Empowered Artificial Intelligence.

\section*{ACKNOWLEDGMENT}

We greatly appreciate the support of this work from the European Commission under the EU Horizon 2020 framework (grant number 101016233), within project PERISCOPE (Pan-European Response to the Impacts of COVID-19 and Future Pandemics and Epidemics), and from the Dutch Research Council (NWO) within the project Designing Rhythms for Social Resilience (grant number 314-99-300).

\section*{AUTHOR CONTRIBUTIONS}

This section uses the Contributor Roles Taxonomy (CRediT); Conceptualization, Y.C.H., T.H.K.H., H.V., A.M., I.N., and A.B.; Methodology, Y.C.H. and T.H.K.H.; Investigation, Y.C.H. and T.H.K.H.; Writing - Original Draft, Y.C.H.; Writing - Review \& Editing, Y.C.H., T.H.K.H., H.V., A.M., I.N., and A.B.; Supervision, A.B.; Project Administration, Y.C.H.; Funding Acquisition, A.B.

\section*{DECLARATION OF INTERESTS}

The authors declare no competing interests.

\section*{DATA AND CODE AVAILABILITY}

This research paper did not use data or code.

\begin{small}

\makeatletter
\renewcommand\@biblabel[1]{#1.}
\let\OLDthebibliography\thebibliography
\renewcommand\thebibliography[1]{
	\OLDthebibliography{#1}
	\setlength{\parskip}{0pt}
	\setlength{\itemsep}{1ex}
}
\interlinepenalty=10000
\renewcommand{\refname}{REFERENCES}
\vspace{0.2ex}
\bibliographystyle{apa-good}
\bibliography{reference}

\begin{thebibliography}{50}
\expandafter\ifx\csname natexlab\endcsname\relax\def\natexlab#1{#1}\fi
\expandafter\ifx\csname url\endcsname\relax
  \def\url#1{{\tt #1}}\fi
\expandafter\ifx\csname urlprefix\endcsname\relax\def\urlprefix{URL }\fi

\bibitem[{Adhane et~al.(2021)Adhane, Dehshibi, and Masip}]{adhane2021deep}
Adhane, G., Dehshibi, M.~M., and Masip, D. (2021).
\newblock {A Deep Convolutional Neural Network for Classification of Aedes
  Albopictus Mosquitoes}.
\newblock {\em {IEEE Access}\/}, {\em 9\/}, 72681--72690.

\bibitem[{Alavi et~al.(2018)Alavi, Verma, Mlynar, and Lalanne}]{alavi2018hide}
Alavi, H.~S., Verma, H., Mlynar, J., and Lalanne, D. (2018).
\newblock {\em {The Hide and Seek of Workspace: Towards Human-Centric
  Sustainable Architecture}\/}, (p. 1–12).
\newblock New York, NY, USA: Association for Computing Machinery.

\bibitem[{Alavi et~al.(2019)Alavi, Verma, Mlynar, and
  Lalanne}]{alavi2019temporality}
Alavi, H.~S., Verma, H., Mlynar, J., and Lalanne, D. (2019).
\newblock {On the temporality of adaptive built environments}.
\newblock In {\em {People, Personal Data and the Built Environment}\/}, (pp.
  13--40). Springer.

\bibitem[{Bennett and Rosner(2019)}]{bennett2019promise}
Bennett, C.~L., and Rosner, D.~K. (2019).
\newblock {The Promise of Empathy: Design, Disability, and Knowing the
  ``Other''}.
\newblock In {\em {Proceedings of the 2019 CHI Conference on Human Factors in
  Computing Systems}\/}, (pp. 1--13).

\bibitem[{Bidwell(2009)}]{bidwell2009community}
Bidwell, D. (2009).
\newblock {Is community-based participatory research postnormal science?}
\newblock {\em {Science, technology, \& human values}\/}, {\em 34\/}(6),
  741--761.

\bibitem[{Bondi et~al.(2021)Bondi, Xu, Acosta-Navas, and
  Killian}]{bondi2021envisioning}
Bondi, E., Xu, L., Acosta-Navas, D., and Killian, J.~A. (2021).
\newblock {Envisioning Communities: A Participatory Approach Towards AI for
  Social Good}.
\newblock In {\em {Proceedings of the 2021 AAAI/ACM Conference on AI, Ethics,
  and Society}\/}, AIES '21, (p. 425–436). New York, NY, USA: Association for
  Computing Machinery.

\bibitem[{Bozzon et~al.(2015)Bozzon, Houtkamp, Kresin, De~Sena, and
  de~Weerdt}]{bozzon2015needs}
Bozzon, A., Houtkamp, J., Kresin, F., De~Sena, N., and de~Weerdt, M. (2015).
\newblock {From Needs to Knowledge: A reference framework for smart citizens
  initiatives}.
\newblock Tech. rep., Amsterdam Institute for Advanced Metropolitan Solutions
  (AMS).

\bibitem[{Brambilla et~al.(2014)Brambilla, Ceri, Mauri, and
  Volonterio}]{brambilla2014community}
Brambilla, M., Ceri, S., Mauri, A., and Volonterio, R. (2014).
\newblock {Community-Based Crowdsourcing}.
\newblock In {\em {Proceedings of the 23rd International Conference on World
  Wide Web}\/}, WWW '14 Companion, (p. 891–896). New York, NY, USA:
  Association for Computing Machinery.

\bibitem[{Brown(1993)}]{brown1993public}
Brown, P. (1993).
\newblock {When the public knows better: Popular epidemiology challenges the
  system}.
\newblock {\em {Environment: Science and Policy for Sustainable
  Development}\/}, {\em 35\/}(8), 16--41.

\bibitem[{Burrell(2016)}]{burrell2016machine}
Burrell, J. (2016).
\newblock {How the machine `thinks': Understanding opacity in machine learning
  algorithms}.
\newblock {\em Big Data \& Society\/}, {\em 3\/}(1), 2053951715622512.

\bibitem[{Carroll et~al.(2019)Carroll, Beck, Boyer, Dhanorkar, and
  Gupta}]{carroll2019empowering}
Carroll, J.~M., Beck, J., Boyer, E.~W., Dhanorkar, S., and Gupta, S. (2019).
\newblock {Empowering Community Water Data Stakeholders}.
\newblock {\em {Interacting with Computers}\/}, {\em 31\/}(3), 492--506.

\bibitem[{Carroll et~al.(2018)Carroll, Beck, Dhanorkar, Binda, Gupta, and
  Zhu}]{carroll2018strengthening}
Carroll, J.~M., Beck, J., Dhanorkar, S., Binda, J., Gupta, S., and Zhu, H.
  (2018).
\newblock {Strengthening community data: towards pervasive participation}.
\newblock In {\em {Proceedings of the 19th Annual International Conference on
  Digital Government Research: Governance in the Data Age}\/}, (pp. 1--9).

\bibitem[{Carroll et~al.(2015)Carroll, Hoffman, Han, and
  Rosson}]{carroll2015reviving}
Carroll, J.~M., Hoffman, B., Han, K., and Rosson, M.~B. (2015).
\newblock {Reviving community networks: hyperlocality and suprathresholding in
  Web 2.0 designs}.
\newblock {\em {Personal and Ubiquitous Computing}\/}, {\em 19\/}(2), 477--491.

\bibitem[{Carton and Ache(2017)}]{carton2017citizen}
Carton, L., and Ache, P. (2017).
\newblock {Citizen-sensor-networks to confront government decision-makers: Two
  lessons from the Netherlands}.
\newblock {\em {Journal of environmental management}\/}, {\em 196\/}, 234--251.

\bibitem[{Ceccaroni et~al.(2019)Ceccaroni, Bibby, Roger, Flemons, Michael,
  Fagan, and Oliver}]{ceccaroni2019opportunities}
Ceccaroni, L., Bibby, J., Roger, E., Flemons, P., Michael, K., Fagan, L., and
  Oliver, J.~L. (2019).
\newblock {Opportunities and risks for citizen science in the age of artificial
  intelligence}.
\newblock {\em {Citizen Science: Theory and Practice}\/}, {\em 4\/}(1).

\bibitem[{Chari et~al.(2017)Chari, Matthews, Blumenthal, Edelman, and
  Jones}]{chari2017promise}
Chari, R., Matthews, L.~J., Blumenthal, M., Edelman, A.~F., and Jones, T.
  (2017).
\newblock {\em {The promise of community citizen science}\/}.
\newblock {RAND}.

\bibitem[{Cheng et~al.(2019)Cheng, Wang, Zhang, O'Connell, Gray, Harper, and
  Zhu}]{cheng2019explaining}
Cheng, H.-F., Wang, R., Zhang, Z., O'Connell, F., Gray, T., Harper, F.~M., and
  Zhu, H. (2019).
\newblock {Explaining decision-making algorithms through UI: Strategies to help
  non-expert stakeholders}.
\newblock In {\em {Proceedings of the 2019 CHI Conference on Human Factors in
  Computing Systems}\/}, (pp. 1--12).

\bibitem[{Doshi-Velez and Kim(2017)}]{doshi2017towards}
Doshi-Velez, F., and Kim, B. (2017).
\newblock {Towards a rigorous science of interpretable machine learning}.
\newblock {\em {arXiv preprint arXiv:1702.08608}\/}.

\bibitem[{Franzen et~al.(2021)Franzen, Kloetzer, Ponti, Trojan, and
  Vicens}]{franzen2021machine}
Franzen, M., Kloetzer, L., Ponti, M., Trojan, J., and Vicens, J. (2021).
\newblock {Machine Learning in Citizen Science: Promises and Implications}.
\newblock {\em {The Science of Citizen Science}\/}, (p. 183).

\bibitem[{Fritz et~al.(2019)Fritz, See, Carlson, Haklay, Oliver, Fraisl,
  Mondardini, Brocklehurst, Shanley, Schade et~al.}]{fritz2019citizen}
Fritz, S., See, L., Carlson, T., Haklay, M.~M., Oliver, J.~L., Fraisl, D.,
  Mondardini, R., Brocklehurst, M., Shanley, L.~A., Schade, S., et~al. (2019).
\newblock {Citizen science and the United Nations sustainable development
  goals}.
\newblock {\em {Nature Sustainability}\/}, {\em 2\/}(10), 922--930.

\bibitem[{Harrington et~al.(2019)Harrington, Erete, and
  Piper}]{harrington2019deconstructing}
Harrington, C., Erete, S., and Piper, A.~M. (2019).
\newblock {Deconstructing community-based collaborative design: Towards more
  equitable participatory design engagements}.
\newblock {\em {Proceedings of the ACM on Human-Computer Interaction}\/}, {\em
  3\/}(CSCW), 1--25.

\bibitem[{Hsu et~al.(2020)Hsu, Cross, Dille, Tasota, Dias, Sargent, Huang, and
  Nourbakhsh}]{hsu2020smell}
Hsu, Y.-C., Cross, J., Dille, P., Tasota, M., Dias, B., Sargent, R., Huang,
  T.-H., and Nourbakhsh, I. (2020).
\newblock {Smell Pittsburgh: Engaging Community Citizen Science for Air
  Quality}.
\newblock {\em {ACM Transactions on Interactive Intelligent Systems (TiiS)}\/},
  {\em 10\/}(4), 1--49.

\bibitem[{Hsu et~al.(2017)Hsu, Dille, Cross, Dias, Sargent, and
  Nourbakhsh}]{hsu2017community}
Hsu, Y.-C., Dille, P., Cross, J., Dias, B., Sargent, R., and Nourbakhsh, I.
  (2017).
\newblock {Community-empowered air quality monitoring system}.
\newblock In {\em {Proceedings of the 2017 CHI Conference on Human Factors in
  Computing Systems}\/}, (pp. 1607--1619).

\bibitem[{Hsu et~al.(2021)Hsu, Huang, Hu, Dille, Prendi, Hoffman, Tsuhlares,
  Pachuta, Sargent, and Nourbakhsh}]{hsu2021project}
Hsu, Y.-C., Huang, T.-H.~K., Hu, T.-Y., Dille, P., Prendi, S., Hoffman, R.,
  Tsuhlares, A., Pachuta, J., Sargent, R., and Nourbakhsh, I. (2021).
\newblock {Project RISE: Recognizing Industrial Smoke Emissions}.
\newblock In {\em {Proceedings of the AAAI Conference on Artificial
  Intelligence}\/}, vol.~35, (pp. 14813--14821).

\bibitem[{Hsu and Nourbakhsh(2020)}]{hsu2020human}
Hsu, Y.-C., and Nourbakhsh, I. (2020).
\newblock {When human-computer interaction meets community citizen science}.
\newblock {\em {Communications of the ACM}\/}, {\em 63\/}(2), 31--34.

\bibitem[{Irwin(2001)}]{irwin2001constructing}
Irwin, A. (2001).
\newblock {Constructing the scientific citizen: science and democracy in the
  biosciences}.
\newblock {\em {Public understanding of science}\/}, {\em 10\/}(1), 1--18.

\bibitem[{Jollymore et~al.(2017)Jollymore, Haines, Satterfield, and
  Johnson}]{jollymore2017citizen}
Jollymore, A., Haines, M.~J., Satterfield, T., and Johnson, M.~S. (2017).
\newblock {Citizen science for water quality monitoring: Data implications of
  citizen perspectives}.
\newblock {\em {Journal of Environmental Management}\/}, {\em 200\/}, 456--467.

\bibitem[{Klein et~al.(2011)Klein, Fatima, McEwen, Moser, Schmidt, and
  Zupan}]{klein2011dismantling}
Klein, P., Fatima, M., McEwen, L., Moser, S.~C., Schmidt, D., and Zupan, S.
  (2011).
\newblock {Dismantling the ivory tower: Engaging geographers in
  university--community partnerships}.
\newblock {\em {Journal of Geography in Higher Education}\/}, {\em 35\/}(3),
  425--444.

\bibitem[{Koekkoek et~al.(2021)Koekkoek, Van~Ham, and
  Kleinhans}]{koekkoek2021unraveling}
Koekkoek, A., Van~Ham, M., and Kleinhans, R. (2021).
\newblock {Unraveling University-Community Engagement: A Literature Review}.
\newblock {\em {Journal of Higher Education Outreach and Engagement}\/}, {\em
  25\/}(1).

\bibitem[{Lotfian et~al.(2021)Lotfian, Ingensand, and
  Brovelli}]{lotfian2021partnership}
Lotfian, M., Ingensand, J., and Brovelli, M.~A. (2021).
\newblock {The Partnership of Citizen Science and Machine Learning: Benefits,
  Risks, and Future Challenges for Engagement, Data Collection, and Data
  Quality}.
\newblock {\em {Sustainability}\/}, {\em 13\/}(14), 8087.

\bibitem[{McClure et~al.(2020)McClure, Sievers, Brown, Buelow, Ditria, Hayes,
  Pearson, Tulloch, Unsworth, and Connolly}]{mcclure2020artificial}
McClure, E.~C., Sievers, M., Brown, C.~J., Buelow, C.~A., Ditria, E.~M., Hayes,
  M.~A., Pearson, R.~M., Tulloch, V.~J., Unsworth, R.~K., and Connolly, R.~M.
  (2020).
\newblock {Artificial intelligence meets citizen science to supercharge
  ecological monitoring}.
\newblock {\em {Patterns}\/}, {\em 1\/}(7), 100109.

\bibitem[{Miller(2019)}]{miller2019explanation}
Miller, T. (2019).
\newblock {Explanation in artificial intelligence: Insights from the social
  sciences}.
\newblock {\em {Artificial intelligence}\/}, {\em 267\/}, 1--38.

\bibitem[{Nickerson(1998)}]{nickerson1998confirmation}
Nickerson, R.~S. (1998).
\newblock {Confirmation bias: A ubiquitous phenomenon in many guises}.
\newblock {\em {Review of general psychology}\/}, {\em 2\/}(2), 175--220.

\bibitem[{Norton et~al.(2012)Norton, Mochon, and Ariely}]{norton2012ikea}
Norton, M.~I., Mochon, D., and Ariely, D. (2012).
\newblock {The IKEA effect: When labor leads to love}.
\newblock {\em {Journal of consumer psychology}\/}, {\em 22\/}(3), 453--460.

\bibitem[{Ottinger(2017)}]{ottinger2017crowdsourcing}
Ottinger, G. (2017).
\newblock {Crowdsourcing undone science}.
\newblock {\em {Engaging Science, Technology, and Society}\/}, {\em 3\/},
  560--574.

\bibitem[{Pataki et~al.(2021)Pataki, Garriga, Eritja, Palmer, Bartumeus, and
  Csabai}]{pataki2021deep}
Pataki, B.~A., Garriga, J., Eritja, R., Palmer, J.~R., Bartumeus, F., and
  Csabai, I. (2021).
\newblock {Deep learning identification for citizen science surveillance of
  tiger mosquitoes}.
\newblock {\em {Scientific reports}\/}, {\em 11\/}(1), 1--12.

\bibitem[{Perell{\'o} et~al.(2021)Perell{\'o}, Cigarini, Vicens, Bonhoure,
  Rojas-Rueda, Nieuwenhuijsen, Cirach, Daher, Targa, and
  Ripoll}]{perello2021large}
Perell{\'o}, J., Cigarini, A., Vicens, J., Bonhoure, I., Rojas-Rueda, D.,
  Nieuwenhuijsen, M.~J., Cirach, M., Daher, C., Targa, J., and Ripoll, A.
  (2021).
\newblock {Large-scale citizen science provides high-resolution nitrogen
  dioxide values and health impact while enhancing community knowledge and
  collective action}.
\newblock {\em {Science of The Total Environment}\/}, {\em 789\/}, 147750.

\bibitem[{Preece et~al.(2019)Preece, Pauw, and Clegg}]{preece2019interaction}
Preece, J., Pauw, D., and Clegg, T. (2019).
\newblock {Interaction design of community-driven environmental projects
  (CDEPs): A case study from the Anacostia Watershed}.
\newblock {\em {Proceedings of the National Academy of Sciences}\/}, {\em
  116\/}(6), 1886--1893.

\bibitem[{Roberts et~al.(2021)Roberts, Driggs, Thorpe, Gilbey, Yeung, Ursprung,
  Aviles-Rivero, Etmann, McCague, Beer et~al.}]{roberts2021common}
Roberts, M., Driggs, D., Thorpe, M., Gilbey, J., Yeung, M., Ursprung, S.,
  Aviles-Rivero, A.~I., Etmann, C., McCague, C., Beer, L., et~al. (2021).
\newblock {Common pitfalls and recommendations for using machine learning to
  detect and prognosticate for COVID-19 using chest radiographs and CT scans}.
\newblock {\em {Nature Machine Intelligence}\/}, {\em 3\/}(3), 199--217.

\bibitem[{Sambasivan et~al.(2021)Sambasivan, Kapania, Highfill, Akrong,
  Paritosh, and Aroyo}]{sambasivan2021everyone}
Sambasivan, N., Kapania, S., Highfill, H., Akrong, D., Paritosh, P., and Aroyo,
  L.~M. (2021).
\newblock {``Everyone wants to do the model work, not the data work'': Data
  Cascades in High-Stakes AI}.
\newblock In {\em {Proceedings of the 2021 CHI Conference on Human Factors in
  Computing Systems}\/}, (pp. 1--15).

\bibitem[{Schneider et~al.(2018)Schneider, Eiband, Ullrich, and
  Butz}]{schneider2018empowerment}
Schneider, H., Eiband, M., Ullrich, D., and Butz, A. (2018).
\newblock {Empowerment in HCI - A survey and framework}.
\newblock In {\em {Proceedings of the 2018 CHI Conference on Human Factors in
  Computing Systems}\/}, (pp. 1--14).

\bibitem[{Shirk et~al.(2012)Shirk, Ballard, Wilderman, Phillips, Wiggins,
  Jordan, McCallie, Minarchek, Lewenstein, Krasny et~al.}]{shirk2012public}
Shirk, J.~L., Ballard, H.~L., Wilderman, C.~C., Phillips, T., Wiggins, A.,
  Jordan, R., McCallie, E., Minarchek, M., Lewenstein, B.~V., Krasny, M.~E.,
  et~al. (2012).
\newblock {Public participation in scientific research: a framework for
  deliberate design}.
\newblock {\em {Ecology and society}\/}, {\em 17\/}(2).

\bibitem[{Shneiderman(2020)}]{shneiderman2020bridging}
Shneiderman, B. (2020).
\newblock {Bridging the gap between ethics and practice: Guidelines for
  reliable, safe, and trustworthy Human-Centered AI systems}.
\newblock {\em {ACM Transactions on Interactive Intelligent Systems (TiiS)}\/},
  {\em 10\/}(4), 1--31.

\bibitem[{Sloan et~al.(2020)Sloan, Moss, Awomolo, and
  Forlano}]{sloan2020participation}
Sloan, M., Moss, E., Awomolo, O., and Forlano, L. (2020).
\newblock {Participation is not a design fix for machine learning}.
\newblock In {\em {Proceedings of the International Conference on Machine
  Learning, Vienna, Austria}\/}.

\bibitem[{Sloane and Moss(2019)}]{sloane2019ai}
Sloane, M., and Moss, E. (2019).
\newblock {AI’s social sciences deficit}.
\newblock {\em {Nature Machine Intelligence}\/}, {\em 1\/}(8), 330--331.

\bibitem[{Susman and Evered(1978)}]{susman1978assessment}
Susman, G.~I., and Evered, R.~D. (1978).
\newblock {An assessment of the scientific merits of action research}.
\newblock {\em {Administrative science quarterly}\/}, (pp. 582--603).

\bibitem[{Wallerstein and Duran(2006)}]{wallerstein2006using}
Wallerstein, N.~B., and Duran, B. (2006).
\newblock {Using community-based participatory research to address health
  disparities}.
\newblock {\em {Health promotion practice}\/}, {\em 7\/}(3), 312--323.

\bibitem[{Wobbrock and Kientz(2016)}]{wobbrock2016research}
Wobbrock, J.~O., and Kientz, J.~A. (2016).
\newblock {Research contributions in human-computer interaction}.
\newblock {\em {Interactions}\/}, {\em 23\/}(3), 38--44.

\bibitem[{Zimmerman et~al.(2007)Zimmerman, Forlizzi, and
  Evenson}]{zimmerman2007research}
Zimmerman, J., Forlizzi, J., and Evenson, S. (2007).
\newblock {Research through Design as a Method for Interaction Design Research
  in HCI}.
\newblock In {\em {Proceedings of the SIGCHI Conference on Human Factors in
  Computing Systems}\/}, CHI '07, (p. 493–502). New York, NY, USA:
  Association for Computing Machinery.

\bibitem[{Zomerdijk and Voss(2010)}]{zomerdijk2010service}
Zomerdijk, L.~G., and Voss, C.~A. (2010).
\newblock {Service design for experience-centric services}.
\newblock {\em {Journal of service research}\/}, {\em 13\/}(1), 67--82.

\end{thebibliography}
\makeatother

\interlinepenalty=4000

\section*{ABOUT THE AUTHORS}

\textbf{Yen-Chia Hsu} is a Postdoctoral Researcher at the Faculty of Industrial Design Engineering, Delft University of Technology. He studies methods to co-design, implement, deploy, and evaluate interactive AI systems that empower communities. He applies crowdsourcing, data visualization, machine learning, computer vision, and data science to engage and assist communities in addressing local environmental and social concerns.

\textbf{Ting-Hao `Kenneth' Huang} is a tenure-track Assistant Professor at the College of Information Sciences and Technology, Pennsylvania State University. His research lies in the intersection of AI and Human-Computer Interaction, imagining new possibilities of human-AI collaborations. He explores the creative and complex domains, like open conversation, writing support, and automatic storytelling, which seem exceptionally challenging to automate. His work aims to move automation beyond low-level, mundane tasks to augment human creativity and sociability.

\textbf{Himanshu Verma} is a tenure-track Assistant Professor at the Faculty of Industrial Design and Engineering, Delft University of Technology. He has a background in HCI, UbiComp and Social Cognition. He is interested in examining collaboration at scale, and his current research is focused on sensing and modeling of interpersonal collaborative processes and how they can be better supported through wearables. In addition, he is also interested in studying the perceptual, cognitive and experiential aspects of human-AI collaboration.

\textbf{Andrea Mauri} is a Postdoctoral Researcher at the Faculty of Industrial Design Engineering, Delft University of Technology. He is also a Research Fellow at the Amsterdam Institute for Advanced Metropolitan Solutions. He has a background in applied machine learning and data science. He is interested in the design, implementation, and evaluation of novel computational methods and tools, focusing on hybrid human-AI methodologies, to support the design processes addressing societal problems by integrating human and societal needs and values.

\textbf{Illah Nourbakhsh} is K\&L Gates Professor of Ethics and Computational Technologies at Carnegie Mellon University, inaugural Executive Director of the Center for Shared Prosperity, and co-director of the Community Robotics, Education and Technology Empowerment Lab. His current research projects explore community-based robotics, including educational and social robotics and ways to use robotic technology to empower individuals and communities.

\textbf{Alessandro Bozzon} is Professor of Human-Centered Artificial Intelligence and Head of the Department of Sustainable Design Engineering, Faculty of Industrial Design Engineering, Delft University of Technology. His research lies at the intersection of human-computer interaction, human computation, user modeling, and machine learning. He is interested in developing methods and tools that support the design, development, control, and operation of AI-enabled systems that are well-situated around actual human characteristics, values, intentions, and behaviors.

\end{small}

\end{document}